\documentclass[10pt,twocolumn,letterpaper]{article}

\usepackage{cvpr}
\usepackage{times}
\usepackage{epsfig}
\usepackage{graphicx}
\usepackage{amsmath}
\usepackage{amssymb}
\usepackage[flushleft]{threeparttable}
\usepackage{color}
\usepackage{multirow}
\usepackage{authblk}
\usepackage[pagebackref=true,breaklinks=true,letterpaper=true,colorlinks,bookmarks=false]{hyperref}

\cvprfinalcopy % *** Uncomment this line for the final submission

\def\cvprPaperID{3098} % *** Enter the CVPR Paper ID here

\ifcvprfinal\fi
\begin{document}
\language0
\lefthyphenmin=2
\righthyphenmin=3
%%%%%%%%% TITLE
\title{Coupled Recurrent Network (CRN)}

\author[1,2,3]{Lin Sun}
\author[4]{Kui Jia}
\author[4]{Yuejia Shen}
\author[2]{Silvio Savarese}
\author[1]{Dit Yan Yeung}
\author[1]{Bertram E. Shi}
\affil[1]{The Hong Kong University of Science and Technology}
\affil[2]{Stanford University} 
\affil[3]{Samsung Strategy and Innovation Center, USA}
\affil[4]{South China University of Technology}
\maketitle
%\thispagestyle{empty}
%%%%%%%%% ABSTRACT
\begin{abstract}

In vision problems, multiple input sources can be used in combination to provide complementary prediction that may be redundant but convey information more effectively. Similarly, many semantic video analysis tasks can benefit from multiple, heterogenous signals. For example, sequences of RGB images and optical flow are usually processed simultaneously to boost the performance of human action recognition in videos. To learn from these heterogenous inputs, existing methods reply two-stream architectures that contain independent, parallel streams of networks. However, two-stream networks do not fully exploit the reciprocal information contained in the multiple signals, let alone exploit it in a recurrent manner. Therefore, we propose, in this paper, a novel recurrent architecture, termed Coupled Recurrent Network (CRN), to deal with multiple input sources. In CRN, the parallel streams of Recurrent Neural Networks (RNNs) are intertwined with each other using Recurrent Interpretation Block (\textit{RIB}) and Recurrent Adaptation Block (\textit{RAB}). \textit{RIB} supports learning of reciprocal representations from multiple signals. \textit{RAB} makes the features adapted to the next recurrence. Different from the training of typical RNNs which stack the loss at each time step or the last time step,  we propose an effective and efficient training strategy for CRN. Experiments show the efficacy of the proposed CRN. In particular, we achieve the new state-of-the-art on the benchmark datasets of human action recognition and multi-person pose estimation.
\end{abstract}

\section{Introduction} \label{SecIntro}
\begin{figure}[htp]
  \centering
  \centerline{\includegraphics[width=5cm]{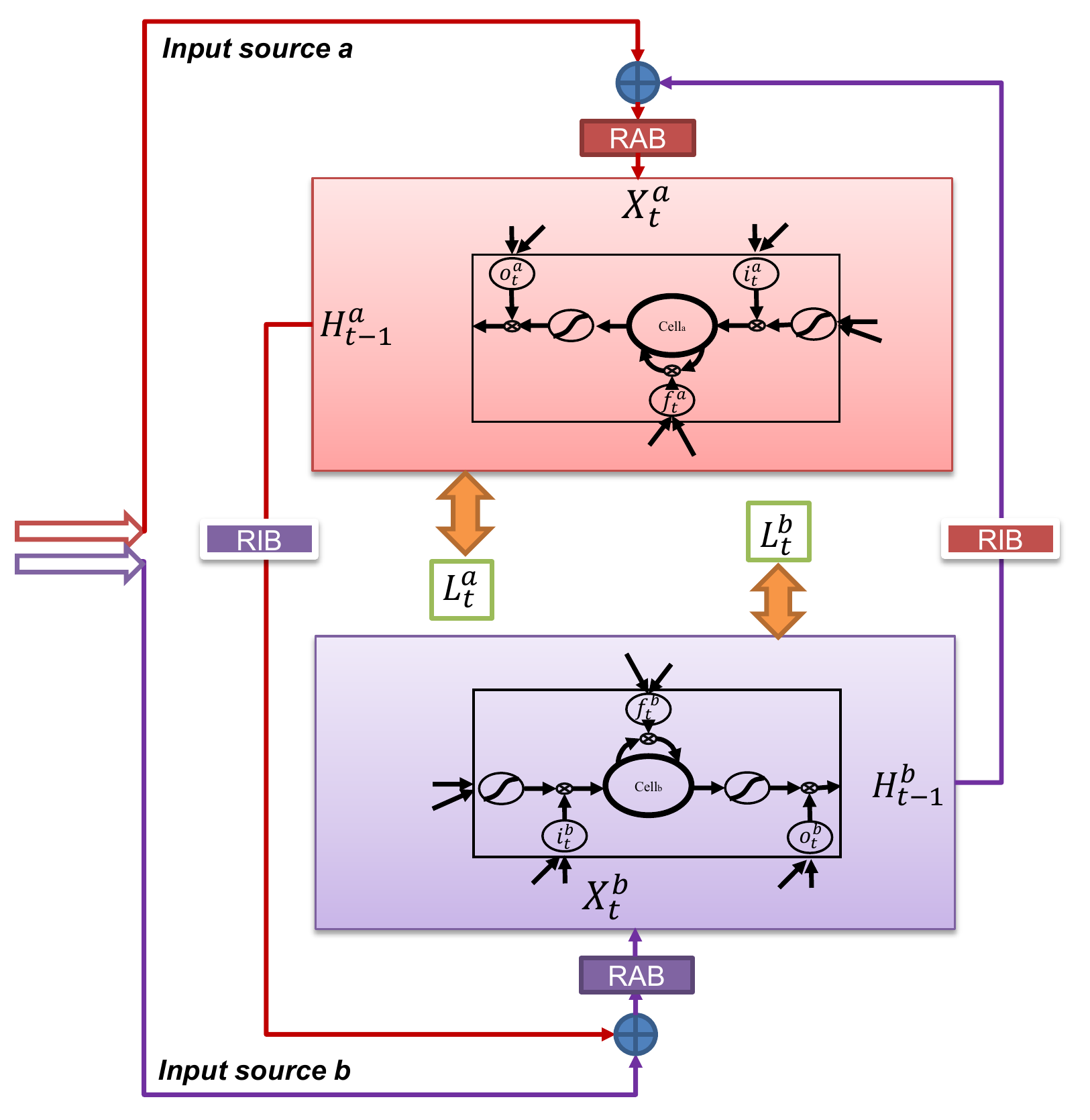}}
  \caption{Illustration of CRN with LSTM units. The blocks outlined in red and purple represent the \textit{Branch A} and \textit{B}. Within each branch, one time-step LSTM is presented, $i_{t}^{*}, f_{t}^{*}$ and $o_{t}^{*}$ are the input gate, forget gate and output gate for the time step $t$, respectively. The filled red and purple rectangles represent the recurrent adaptation block (\textit{RAB}). The filled bright red and purple rectangles represent recurrent interpretation block (\textit{RIB}). } \label{fig:crn-lstm}
\end{figure}

Many computer vision tasks rely on semantic analysis of data in sequential forms. Typical examples include video-based human action recognition \cite{jeff2015lrcn}, image/video captioning \cite{yu2015caption}, speech recognition \cite{graves2013speech} etc. In other cases, tasks of interest might be recast as sequential learning problems, so that their learning objectives can be easily and iteratively achieved. For example, in human pose estimation, the joint locations can be predicted using multi-stage CNNs, the hidden features produced by one stage are used as input for the next stage. This multi-stage scheme for pose estimation can be cast using the recurrent scheme that the hidden output from one time step will be fed into in the next time step for refinement. 

RNNs in which each neuron or unit can use its internal memory to maintain information of the previous input have become the de facto learning models for many computer vision tasks whose observed input data are or can be recast in a sequential form. In these tasks, we might have access to multiple and heterogenous input sources. In general, having multiple input sources working together towards the same goal, can contribute to a more robust algorithm. The input sources each may be more effective at expressing one form or aspect of information than others. This is the reason why the two-stream architectures is widely accepted in human action recognition. RGB frames and corresponding motion inputs (e.g. optical flow or RGB difference) is used as multiple input sources which, separately, will pass through the independent network. The results of the individual models will be combined together to predict the actions happened in the video. 
%For example, in human action recognition task, a sequence of RGB frames and optical flows will be fed into two independent models. The output probabilities from these two models will be averaged or weightedly averaged. The predicted category will be the largest entry of the fused probability. 

The independent use of networks on different input sources and fusing the probabilities at the end does not fully exploit the reciprocal information between each other. Moreover, there are few researches on how to exploit reciprocal information between complementary inputs. Therefore, in this paper, we propose an end-to-end architecture, Coupled Recurrent Network (CRN) to investigate how to aggregate the features to better fuse spatial and temporal information and achieve more effective sequential learning from multiple inputs. CRN has two recurrent branches, i.e. \textit{branch A} and \textit{branch B}, each of which takes one input source. Different from two-stream architecture which is trained independently, during the learning of CRN, \textit{branch A} and \textit{branch B} can `\textit{communicate}' with each other. In order to well interpret the reciprocal information between them, within CRN, we propose two modules, recurrent interpretation block (\textit{RIB}) and recurrent adaptation block (\textit{RAB}). The hidden output from \textit{branch B} will pass through \textit{RIB} to extract reciprocal representations for \textit{branch A}. The extracted reciprocal information will be concatenated with current input sources of \textit{branch A} and then they will pass through the \textit{RAB} to obtain the re-mapped input for \textit{branch A}. The distilling module \textit{RIB} and remapping module \textit{RAB} is shared inside the recurrent networks. An illustration of a CRN and its corresponding computational flow is shown in Fig. \ref{fig:crn-lstm}. 

%($[X_{1}^{a},X_{2}^{a}, \cdots X_{T}^{a}]$ and $[X_{1}^{b},X_{2}^{b}, \cdots X_{T}^{b}]$) simultaneously. $[H_{1}^{a},H_{2}^{a}, \cdots H_{T}^{a}]$ and $[H_{1}^{b},H_{2}^{b}, \cdots H_{T}^{b}]$ are sequential hidden output from each branch. 

The proposed CRN can be generlized to many computer vision tasks which can be understood by two or several input sources. In this paper, we apply the proposed CRN to two human centric problems, i.e. human action recognition and multi-person pose estimation. In human action recognition, a sequence of RGB frames and corresponding motion signals (e.g. optical flows or RGB difference) are the two heterogenous input sources for two branches in CRN. Two cross entropy losses for the same recognition target with identical form are applied at the end of each branch. While in multi-person pose estimation, besides the commmonly used individual body joints, a field of part/joint affinities that characterizes pair-wise relations between body joints \cite{cao2016realtime} is used as the additional supervision information. Two different regression losses, one for joint estimation and the other one for vector prediction, are applied at the end of each network. 

The standard procedure for training a RNN is to apply appropriate loss for each input at each time step or only apply at the last time step. However, in the experiments, we find neither of two training strategies work well for CRN. Having supervision at each time step seems to make supervision signals assertive and arbitrary and the whole training becomes numerically unstable, leading to a poor performance; while having supervision only at the last time step makes supervision signal too weak to reach the end and leads to the performance drop on the considered tasks. Therefore, we propose a new training scheme which has a good balance of the supervision strength along time steps. Apart from having the loss at the last time step, we randomly select some previous time steps for supervision. Only the losses at selected time step will contribute to the back-propagation.

Comparative experiments show that our proposed CRN outperforms the baselines by a large margin. CRN sets a new state-of-the-art on benchmark datasets of human action recognition (e.g., HMDB-51 \cite{kuehne2011hmdb51}, UCF-101 \cite{khurra2012ucf101} and larger dataset, Moments in Time \cite{monfort2017moments}) and multi-person pose estimation (e.g., MPII \cite{andriluka14cvpr}). Moreover, since better reciprocal information can be exploited within CRN, using RGB and RGB differences as input sources, CRN can achieve more than $93\%$ accuracy on the UCF-101. Compared to optical flow, RGB difference can be calculated online without burden. Therefore, more than 200 FPS can be achieved for real-time action recognition. We summarize the contributions as below:

\begin{itemize}
\item We propose a novel architecture, Coupled Recurrent Network (CRN), to deal with multiple input sources in a reciprocal and recurrent manner. In order to interpret the representations of each input source for the other in a recurrent approach, recurrent interpretation block (\textit{RIB}) and recurrent adaptation block (\textit{RAB}) are proposed. \textit{RIB} is used to distill the useful information from the output of one branch for the other branch. \textit{RAB} provides the remapping of two concatenated representations.  

\item Two tasks, i.e. human action recognition and human pose estimation, are investigated and analyzed using proposed CRN. However, our proposed method can be generalized to many computer vision tasks which initially is or can be recast as sequential learning. 

\item A effective and efficient training strategy for CRN is proposed. 

\item The state-of-the-art performance for real-time action recognition can be achieved using CRN which will definitely push forward the application of action recognition in industry. 

\item Extensive quantitative and qualitative evaluations are presented to analyze and verify the effectiveness of our proposed method. We also conduct several ablation studies to validate our core contributions. 
\end{itemize}

\section{Related Works}

In this section, we first provide a brief review of using RNNs for different computer vision tasks, particularly focusing on those algorithms that deal with multiple sources of sequential inputs. Then, we review representative methods for action recognition and human pose estimation. 

\subsection{RNNs for Multiple Sources of Sequential Data}

RNN is a class of artificial neural network where connections between nodes form a directed graph along a temporal sequence. Therefore, RNNs natually have advantages in processing sequential data. Among all of recurrent networks, Long Short-Term Memory network, a.k.a LSTM \cite{hochreiter1997lstm}, has been observed to be the most effective. On one hand, with sequential inputs, RNNs can produce outputs that are sequences.  Countless learning tasks require dealing with this situation, such as image captioning \cite{mao2014deep}, speech recognition \cite{Graves05framewisephoneme}, and musicial composition \cite{eckmusic2002}. On the other hand, RNNs can provide the prediction from a sequential input, such as handwriting recognition \cite{graves2009handwriting}, video analysis \cite{sub2017video} etc.  RNNs also extends its success to sequential data with multiple modalities, e.g. RNNs integrating the video and audio information for video captioning \cite{wang2018caption}. 

\subsection{Human Action Recognition}
%Data, particularly images, is the fuels that equate these models to rocket engines. 
Neural networks have achieved remarkable results for image-based vision tasks, so, not surprisingly, there have been many recent attempts to extend these successes to videos. However, human actions in videos are three-dimensional~(3D) spatio-temporal signals, therefore, typical 2D networks can not handle it well. For that reason, \cite{simonyan2014twostream} proposes to use a two-stream architecture. In this design, spatial network is processing RGB frames and temporal network is processing optical flows, the probability will be averaged or weightdely averaged at the end. Multiple input sources can be used in combination to provide complementary methods that may be redundant but convey information more effectively. Here, RGB images provide apperance representations, while optical flows explicitly capture motion information of videos. Many works follow the two-stream architecture. Tran et al. explore 3D ConvNets \cite{du2015C3D} on realistic and large-scale video datasets, where they try to learn both appearance and motion features with 3D convolution operations. Sun \cite{sun2015fstcn} propose a factorized spatiotemporal ConvNet using the difference between neighobouring RGB images as additional motion information. Wang et.al \cite{linmin16tsn} propose a temporal segment network (TSN), which is based on the idea of long-range temporal structure modeling, for RGB images and optical flows. Besides the original two stream networks, \cite{yudistira2017} applies a gating CNN to combine the information from RGB images and optical flows. \cite{park2016fusion} presents new approaches to combine different sources of knowledge in spatial and temporal networks. They propose feature amplification, where they use an auxiliary, hand-crafted, feature (e.g. optical flow) to perform spatially varying soft-gating on intermediate CNN feature maps. They present a spatially varying multiplicative fusion method for combining multiple CNNs trained on different sources. Even the algorithm is sophisticated, the performance is far from satisfying. \cite{feichtenhofer16fusion} proposes that spatio-temporal features can be fused at a convolution layer. In \cite{dynamic2017}, the authors extend the residual block to temporal domain within each stream. This modification makes their algorithm similar to RNNs, that is, the information from the current step will be fed into the next step. \cite{i3d2017} inflates the 2D kernels pre-trained on ImageNet to 3D for videos. 

RNNs is another choice for this task. \cite{jeff2015lrcn, yuehei2015twolstm} proposed to train video recognition models using LSTMs that capture temporal state dependencies and explicitly model short snippets of ConvNet activations. Ng et al. \cite{yuehei2015twolstm} demonstrated that two-stream LSTMs outperform improved dense trajectories (iDT) \cite{wang2013idt} and two-stream CNNs \cite{simonyan2014twostream}, although they need to pre-train their architecture on extra videos. VideoLSTM \cite{li16videolstm} applies convolutional operations within LSTM on sequences of images or feature maps. Additionally, an attention model is stacked on top of the ConvLSTM to further refine the temporal features. Sun et.al \cite{lin2017l2stm} also propose a lattice LSTM for the long and complex temporal modeling. These two-stream RNNs are all trained independently and combined on the probability level. Even lattice LSTM has joint training on the gates between the two streams, their representations are not completely coupled. 

\subsection{Human Pose Estimation}

Human pose estimation considers to use multiple branches to improve the precision as well. Based on the multi-stage work       \cite{wei2016cpm} , Cao et al. \cite{cao2016realtime} present a real-time pose estimation method by adding a bottom-up representation of association scores via part affinity fields (PAFs). Joint associate network works as backup when joint detection network fails in some context, a robust multi-person pose estimation is provided. 

\section{LSTM-based Coupled Recurrent Network}\label{SecAlgo}
LSTM is commonly used recurrent architecture, due to its superior performance in many tasks. In this paper, we adopt LSTM, particularly convolutional LSTM (ConvLSTM) \cite{shi2015convlstm} as the basic unit within CRN. To simplify presentation, in the rest of the paper, the abbreviation LSTM instead of ConvLSTM will be used.
%Usually, an LSTM contains a cell memory to remember, an input, a forget and an output gate to control.

Formally, let's denote $\{ {X}^{a}_{t},  {X}^{b}_{t}\}$ as two input sources at the time step $t$, where $a$ and $b$ are the indices of input sources. $\{ {X}^{a}_{t},  {X}^{b}_{t}\}$ are usually in the form of 2D images or feature representations. A CRN contains two branches, each of which handles one input source. Since \textbf{two branches in CRN are symmetric}, in the following paragraph, we will illustrate \textit{branch A} step by step to present the whole flow of computation. Starting from the cell memory $\widetilde{C}^{a}_t$ at time $t$ which maintains the information over time within recurrence, 

\begin{eqnarray}\label{eqn:memorycell}
\begin{aligned}
\widetilde{C}^{a}_t &= tanh(W_{xc} \ast {\hat{{X}}^{a}_{t}} + W_{hc} \ast {H}^{a}_{t-1}),\\
\end{aligned}
\end{eqnarray}
where $W_{xc}$ and $W_{hc}$ are, respectively, the weights for the input and hidden states. The symbol $\ast$ denotes the convolution operation and ${H}^{a}_{t-1}$ is the hidden output at time step $t-1$. The concatenation of ${X}_t^{a}$ and the interpreted reciprocal hidden representations $\hat{{H}}^{b}_{t-1}$ (described in  \ref{sec:rib}) from \textit{branch B} will pass through \textit{Recurrent Adapted Block (RAB)} to obtain the remapped representations ${\hat{{X}}}^{a}_{t}$ :
\begin{eqnarray}\label{eqn:rab}
\begin{aligned}
{\hat{{X}}}^{a}_{t} = {\cal{R}}^a_{ab}({X}^{a}_t \oplus {\hat{H}}^{b}_{t-1}) ,
\end{aligned}
\end{eqnarray}
where ${\cal{R}}^a_{ab}$ denotes the functions of \textit{RAB}. ${\cal{R}}^a_{ab}$ can be one or several convolutional layers which are shared at different time steps. $\oplus$ is the concatenation operation. 

The input gate $i^{a}_t$ and forget gate $f^{a}_t$ at time step $t$ are computed using remapped ${\hat{{X}}}^{a}_{t}$ as
\begin{eqnarray}\label{eqn:gates}
\begin{aligned}
i^{a}_t &= \sigma(W_{xi}\ast {\hat{{X}}}^{a}_{t} + W_{hi}\ast {H}^{a}_{t-1}), \\
f^{a}_t &= \sigma(W_{xf}\ast {\hat{{X}}}^{a}_{t} + W_{hf}\ast {H}^{a}_{t-1}), \\
\end{aligned}
\end{eqnarray}
where $W_{x.}, W_{h.}$ are distinct weights for the input and hidden states in gates. At the end of time step $t$, we can obtain the updated memory cell $C^{a}_t$ from the previous memory cell $C^{a}_{t-1}$ and $\widetilde{C}_t^{a}$:
\begin{eqnarray}\label{eqn:memory}
\begin{aligned}
C^{a}_t &= f^{a}_t\circ C^{a}_{t-1} + i^{a}_t\circ \widetilde{C}^{a}_t,\\
\end{aligned}
\end{eqnarray}
where `$\circ$' denotes the pixel-wise multiplication, a.k.a, the Hadamard product. The input and forget gate together determine the amount of dynamic information entering/leaving the memory cell. The final hidden output ${H}^{a}_t$  is controlled by output gate $o^{a}_t$, 
\begin{eqnarray}\label{eqn:convLSTM}
\begin{aligned}
o^{a}_t &= \sigma(W_{xo}\ast {\hat{{X}}}^{a}_{t} + W_{ho}\ast {H}^{a}_{t-1}), \\
{H}^{a}_t &= o^{a}_t\circ tanh(C^{a}_t).
\end{aligned}
\end{eqnarray}
where $W_{xo}, W_{ho}$ are distinct weights for the output gate. The \textit{branch B} is processed in a similar way to obtain ${H}^{b}_t $. Although the features within CRN are coupled, the parameters are initialized independently for each branch. 

\subsection{Adapting to Different Tasks}
After obtaining ${H}^{a}_t$, we stack additional transformation layer(s) (e.g., convolutional or linear) to extract final representations ${Q}^{a}_{t}$, adapting to different supervision tasks. 
\begin{equation}\label{equ:loss}
{Q}^{a}_{t} = {{G}}^{a}(H^{a}_t) ,
\end{equation}
where ${{G}}^{a}$ denotes the transformation function(s). 

In the training process, identical or different losses can be applied to each branch. In our case, ${\cal{L}}^a(\cdot)$ and ${\cal{L}}^b(\cdot)$ are two loss functions, 
\begin{eqnarray}\label{eqn:loss}
\begin{aligned}
L^a_t &= {\cal{L}}^a({Q}^{a}_{t}, Q_{t}^{*a}),  \\
L^b_t &= {\cal{L}}^b({Q}^{b}_{t}, Q_{t}^{*b}), 
\end{aligned}
\end{eqnarray}

where $Q_{t}^{a*}$ and $Q_{t}^{b*}$ is the supervision target for \textit{branch A} and \textit{branch B}. Depending on the tasks, ${\cal{L}}^a(\cdot)$ and ${\cal{L}}^b(\cdot)$ can be the euclidean distance or cross-entropy loss etc. $L^a_t$ and $L^b_t$ are the losses for \textit{branch A} and \textit{B} at time step $t$, respectively. The overall loss for CRN is 
\begin{equation}\label{equ:loss}
L = \sum_{t\in {\cal{T}}} (L_t^a + L_t^b) .
\end{equation}
where ${\cal{T}}$ is a set of selected time steps for supervision. 

\subsection{Extracting Reciprocal Representations} \label{sec:rib}
\begin{figure}[htp]
  \centering
  \centerline{\includegraphics[width=7cm]{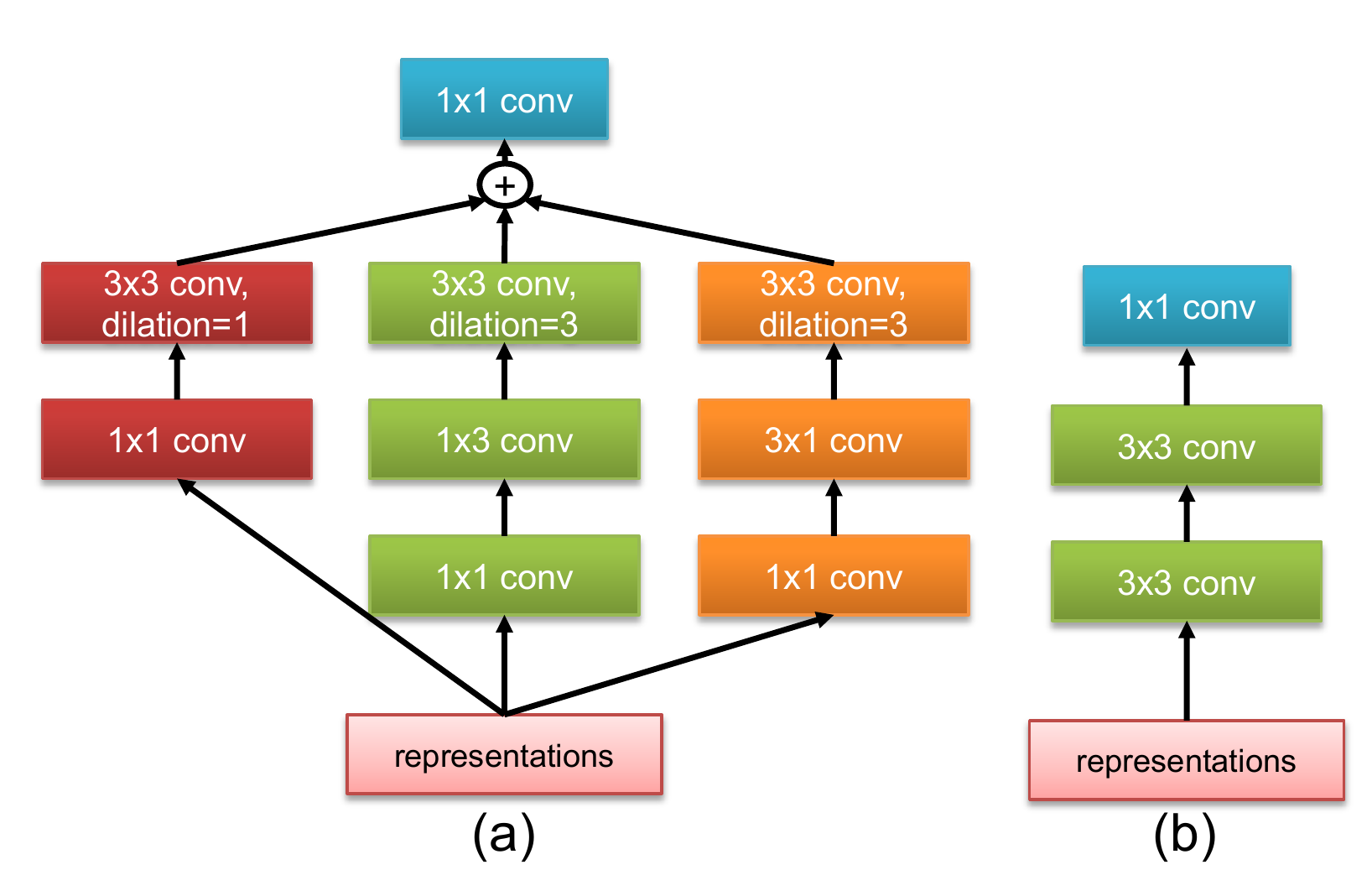}}
  \caption{Illustration of two architectures of Recurrent Interpretation Block (\textit{RIB}). (a) \textit{RIB} (b) \textit{sRIB} } \label{fig:rib}
\end{figure}

The interpreted reciprocal representation $\hat{{H}}^{b}_{t-1}$ can be obtained by passing the hidden output ${H}^{b}_{t-1}$ from \textit{branch B} through the \textit{Recurrent Interpretation Block (RIB)}:

\begin{eqnarray}\label{eqn:rib}
\begin{aligned}
{\hat{{H}}}^{b}_{t-1} = {\cal{R}}^a_{ib}({H}^{b}_{t-1}) .
\end{aligned}
\end{eqnarray}
where ${\cal{R}}^a_{ib}$ denotes the \textit{RIB}. Like \textit{RAB}, it consists of one or several convolutional layers which hold shared parameters at different time steps. Although the input sources or the supervision targets of two branches are related, directly concatenating the input sources (i.e. $X^a_{t}$ or $X^b_{t}$) with the hidden output (i.e. $H^b_{t}$ or $H^a_{t}$) leads to little improvement. Certain input source can provide a richer context information, the key is how to `borrow' the really useful and complementary information from the other. Therefore, \textit{RIB} is designed to distill the reciprocal information from each other at every time step. As illustrated in Fig. \ref{fig:rib} (a), in order to effectively extract the reciprocal information, \textit{RIB} has a similar design like an inception module which has three parallel convolutions with different dilation ratios. To investigate the importance of capacity of \textit{RIB}, a simplized version, \textit{sRIB} is provided in Fig. \ref{fig:rib} (b). Experiments present that neither direct concatenation nor \textit{sRIB} perform as effectively as \textit{RIB}. It indicates that \textit{RIB} is a useful design for CRN and when learning transferable knowledge from different input sources, more complicated architecture can distill more reciprocal information. 

%To train our CRN effectively, Eq. \ref{eqn:loss} is only applied to a selected set ${\cal{T}}$ of time steps. ${\cal{T}}$ consists of the last time step $T$ and randomly selected one or several time steps from $1$ to $T-1$
\begin{figure*}
  \centering
  \centerline{\includegraphics[width=15cm, height=8cm]{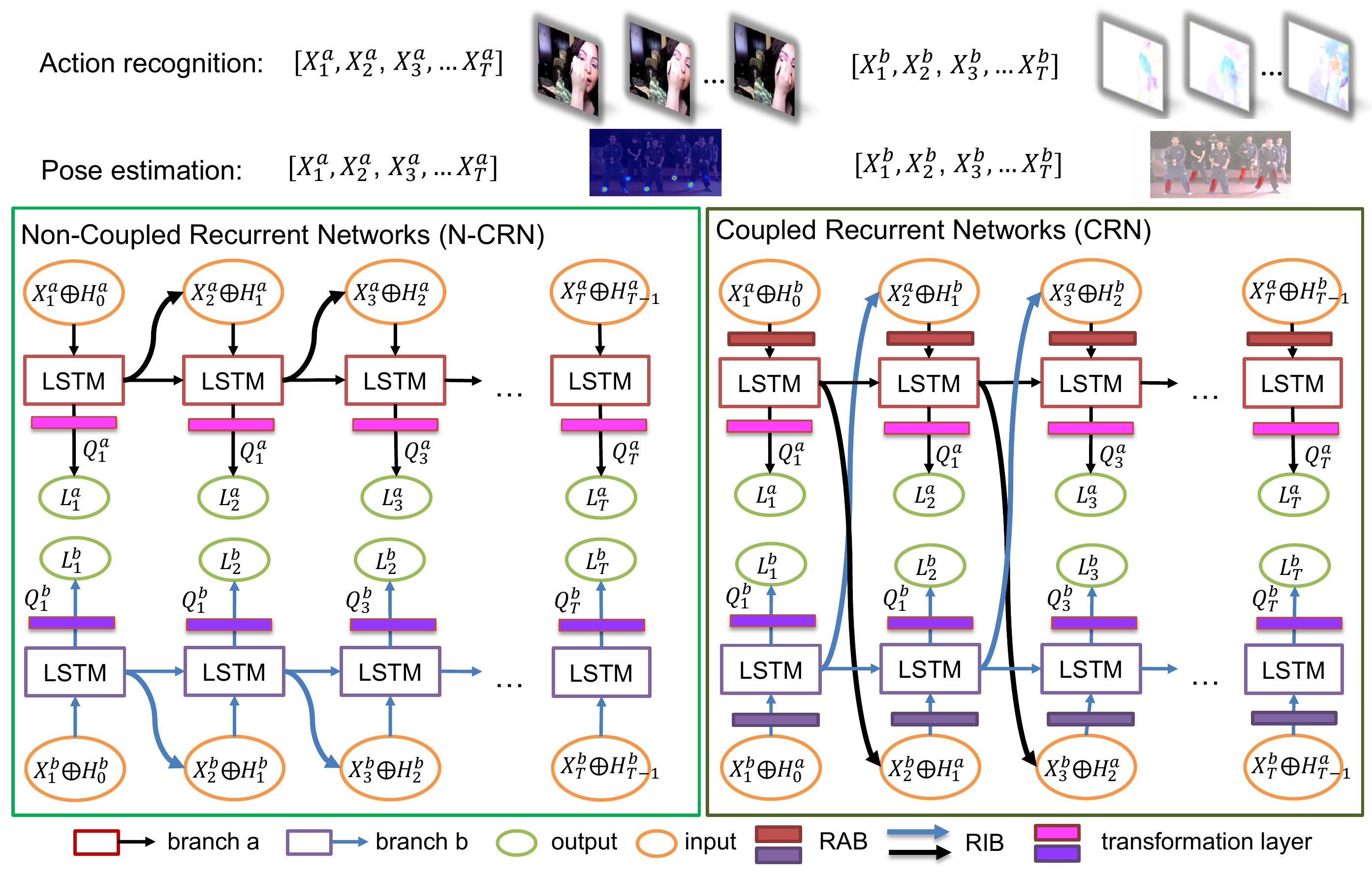}}
  \caption{ Illustration of CRNs for specific tasks. Each block represents one time step of recurrent networks. Blocks of different colors process different input sources. Left (light green) presents Non-Coupled Recurrent Network (N-CRN). Right (dark green) shows the architecture of CRN.  $[X_{1}^{a},X_{2}^{a}, \cdots X_{T}^{a}]$ represent input source $a$ and $[X_{1}^{b},X_{2}^{b}, \cdots X_{T}^{b}]$ is for the source $b$. $[\hat{H}_{1}^{a},\hat{H}_{2}^{a}, \cdots \hat{H}_{T}^{a}]$, $[\hat{H}_{1}^{b},\hat{H}_{2}^{b}, \cdots \hat{H}_{T}^{b}]$ are the interpreated hidden for each source, respectively. The inputs for the human action recognition and human pose estimation are different and shown on the top. Rectangles with filled colors denote \textit{RAB} and thick arrow lines illustrate \textit{RIB}. Rectangles with filled light red and light purple represents transformation layers. Better viewed in color.} \label{fig:framework-pose}
\end{figure*}

\section{Use of CRN for Specific Tasks}\label{SecUseCRNs}
We present, in this section, how can our proposed CRNs be applied to two tasks, namely video based human action recognition and image based multi-person pose estimation. Fig. \ref{fig:framework-pose} gives an illustration, where recurrent networks are unrolled to better present the processing flow. The architecture of non-coupled recurrent network (N-CRN) is provided on the left side of the figure as a baseline while the exemplar design of CRN is shown on the right side. N-CRN is just a modified two-stream architecture in which input source concatenate the previous hidden output of itself independently. $[X_{1}^{*},X_{2}^{*}, \cdots X_{T}^{*}]$ represents the sequential data for input source $a$ and $b$ where `*' can be $a$ or $b$. The inputs can be the raw images or feature maps extracted from intermediate layers. $[\hat{H}_{1}^{*},\hat{H}_{2}^{*}, \cdots \hat{H}_{T}^{*}]$ are interpreted hidden outputs,  $[Q_{1}^{*},Q_{2}^{*}, \cdots Q_{T}^{*}]$ represent the final representations from each branch. 

In human action recognition, the input is a sequence of RGB frames and a corresponding sequence of optical flows or RGB differences, and the objective is to classify the input video as one of the action categories. The output of \textit{branch A} at time step $t$ is $Q_{t}^{a} \in \Omega^{1\times N}$, $N$ is the number of categories. $Q_{t}^b$ has the same dimension and supervision target as $Q_{t}^a$. Two cross entropy losses with identical form, are added at the end of each input source. Two probabilities from two branches will be averaged for final prediction.  

In multi-person pose estimation, the input is a sequence of the repeated images, and the objective is to estimate 2D locations of body joints for each person in the image. CRN simultaneously outputs a set of hidden features $Q_{t}^{a}$ for heat maps and $Q_{t}^{b}$ of for 2D PAFs prediction, which encode the degrees of association between body joints. $Q_{t}^{a} \in \Omega^{w\times h \times M}$ has $M$ feature maps with $w \times h$ resolution and each of them corresponds to one body joint at time step $t$. $Q_{t}^{b}\in \Omega^{w\times h \times 2N}$ has $N$ vectors whose width is $w$ and height is $h$. Each of vector corresponds to a limb of the human body at the time step $t$. $l2$ loss is applied at the end of each branch to minimize the heat maps and PAFs. We follow the greedy relaxation as \cite{cao2016realtime} to find the optimal parsing. 

\section{Experiments}

We apply our proposed CRNs to two computer vision tasks. For human action recognition, we use three large scale benchmark datasets:

\textbf{UCF-101} \cite{khurra2012ucf101} is composed of realistic web videos. It has 101 categories of human actions with more than 13K videos. It has three split settings to separate the dataset into training and testing videos. The mean classification accuracy over these three splits is used for evaluation. 

\textbf{HMDB-51} \cite{kuehne2011hmdb51} has a total of 6766 videos organized as 51 distinct action categories. Similar to UCF-101, HMDB-51 has three split settings, and the mean classification accuracy over these three splits is used for evaluation.

\textbf{Moments in Time} \cite{monfort2017moments} consists of over 1,000,000 3-second videos corresponding to 339 different verbs depicting an action or activity. Each verb is associated with over 1,000 videos, resulting in a large balanced dataset for learning a basis of dynamical events from videos.  

For human pose estimation, we use a multi-person pose estimation : 

\textbf{MPII dataset} \cite{andriluka14cvpr} consists of 3844 training and 1758 testing groups with crowded, occlusion, scale variation and overlapped people from the real world. We use 3544 images for training, leaving 300 images for validation.

\subsection{Implementation Details}
For the action recognition, bninception \cite{ioffe2015bn} and inceptionv3 \cite{inceptionv316szegedy} are used as backbone networks,  features from the last convolutional layers will be fed into the CRN. RGB frames and corresponding optical flows, and RGB frames and RGB differences will be paired and pass through these backbone simultaneously. For each branch in CRN, it is a two-layer LSTM. Within convolutions, all the kernels are $3\times3$  and the number of hidden feature maps is 512. Batches of ten sequential frames will be fed into system for training. The detailed architecture of \textit{RIB} is shown in Fig. \ref{fig:rib} (a) and the \textit{RAB} is just one convolutional layer with kernel size $1 \times 1$. In the experiments, we find adding more layers for \textit{RAB} or make it complicated does not help the final performance. The transformation layer $G$ is a global pooling followed by a fully connection layer. The whole system can be trained end-to-end using SGD. The initial learning rate for CRN is 1e-2 and for backbone is 1e-3. The momentum is set to 0.9 and weight decay is 5e-4. Besides the last time step, we randomly select additional one time step from the previous time steps for back-propagation. When testing, we regularly sample four clips and average their probabilities. 

Like \cite{cao2016realtime}, we pre-process pose estimation images using VGG-19 \cite{simonyan14vgg} which is pre-trained on ImageNet \cite{russakovsky2015imagenet}. The whole system is trained using SGD where the initial learning rate for CRN and backbone is 2e-4 and 5e-5, respectively. For each branch in CRN, a two-layer LSTM with all $7\times 7$ convolutions is applied. Batches of ten repeated images will be fed into system. We use the same architecture of \textit{RIB} and \textit{RAB} as described in the action recognition task.

All the implementations are built using pytorch \cite{pytorch}. All the experiments are run on GTX 1080 and evaluated under the same settings as instructed. 

\begin{table}[htb] \caption{Evaluation of $RIB$ on UCF-101 and HMDB-51 under different settings} \label{t:rib}
\setlength{\tabcolsep}{6.5pt}
\renewcommand{\arraystretch}{1.2}
\scriptsize
\centering
\begin{tabular}{c|c|c|c|c|c} \hline
\multicolumn{6}{c}{UCF-101} \\
\hline
\multirow{2}{*}{No.} & \multirow{2}{*}{Training settings} & \multicolumn{2}{c|}{RGB+Flow} & \multicolumn{2}{c}{RGB+Diff} \\ \cline{3-6}
{}& {}&S-Nets&T-Nets &S-Nets&T-Nets \\ 
\hline
%1 &CRN (vgg16 + \textit{sRIB}) &85.9\% &87.8\% &82.8\% & 83.3\% \\
%2 &CRN (vgg16 + \textit{RIB}) &86.4\% &88.8\% &84.8\% & 86.2\%  \\
1 &CRN (bninception) &88.3\% &90.6\% &86.7\% & 87.9\% \\
2 &CRN (bninception + \textit{sRIB}) &90.4\% &92.3\% &88.7\% & 89.3\% \\
3 &CRN (bninception + \textit{RIB}) &91.4\% &93.0\% &89.5\% & 90.7\% \\
4 &CRN (inceptionv3) &91.0\% &92.3\% &88.9\% & 90.2\% \\
5 &CRN (inceptionv3 + \textit{sRIB}) &91.8\% &92.9\% &90.4\% &90.8\% \\
6 &CRN (inceptionv3 + \textit{RIB}) &93.0\% &93.5\% &91.2\% &91.6\% \\
\hline
\multicolumn{6}{c}{HMDB-51} \\
\hline
%1 & CRN (vgg16 + \textit{sRIB}) &52.9\% &58.0\% &46.8\% &48.5\%\\
%2 & CRN (vgg16 + \textit{RIB}) &53.2\% &59.6\% &48.7\% &51.2\% \\
1 & CRN (bninception) &54.7\% &61.8\% &52.7\% &54.9\% \\
2 & CRN (bninception + \textit{sRIB}) &58.6\% &63.5\% &55.4\% &57.6\% \\
3 & CRN (bninception + \textit{RIB}) &60.3\% &67.5\% &55.9\% &59.0\%\\
4 & CRN (inceptionv3) &61.5\% &64.1\% &57.1\% &58.2\% \\
5 & CRN (inceptionv3 + \textit{sRIB}) &63.1\% &65.8\% &59.0\% &59.9\% \\
6 & CRN (inceptionv3 + \textit{RIB}) &64.4\% &67.7\% &60.9\% &60.9\% \\
\hline
\end{tabular}
\end{table}

\subsection{Evaluation on Action Recognition}
\textbf{The effect of \textit{RIB}}: In order to verify that distilling reciprocal information from complementary input source is useful, we evaluate CRN using different backbones with/without different \textit{RIB} architectures. The detailed results are presented in Table \ref{t:rib}, where 'S-Nets` denotes the spatial networks and 'T-Nets` denotes the temporal networks. The paired input sources can be RGB frames and optical flows or RGB frames and corresponding RGB differences. From the table, 1 vs. 2, 3 and 4 vs. 5, 6 in UCF-101 and HMDB-51, respectively, compared to the results generated with  \textit{RIB}, directly concatenating the hidden output from the other branch without distilling/\textit{RIB} performs much worse. And under all the backbones, CRN with \textit{RIB} can achieve better performance, with about $1\%$ performance gain, over CRN with \textit{sRIB}. Stronger \textit{RIB} module makes our CRN distill more appropriate representations to the other branch. All the results indicate that how and how much information is distilled from complementary input source affects the final performance. What is more, the whole procedure is recurrent, iteratively refining the interpreted representations makes our coupled learning generate better representations for each input source. A CRN built on the inceptionv3 backbone and with \textit{RIB} can achieve the best performance. Without specific notation, in the following paragraph, we adopt this architecture for comparison. Surprisingly, the performance of CRN achieved using one branch on two benchmark datasets is already better than fused performance of some sophisticated two-stream algorithms. 

\begingroup
\setlength{\tabcolsep}{5.0pt}
\renewcommand{\arraystretch}{1.0}
\begin{table}[htb]\caption{Evaluation with training strategies on split 1 of HMDB-51 } \label{t:training-sample}
\centering
\scriptsize
\begin{tabular}{c|c|c|c|c} \hline
Strategy & setting & S-Nets & T-Nets (Flow) & T-Nets (Diff)  \\
\hline
a & CRN (bninception) & 55.5\% & 62.2\% & 56.4\%\\
b & CRN (bninception) & 47.3\% & 57.2\% & 49.7\%\\
c & CRN (bninception) & \textbf{58.1}\% & \textbf{67.5}\% & \textbf{59.0}\% \\
d & CRN (bninception) & 56.9\% & 66.9\% & 58.6\%\\
\hline
\end{tabular}
\end{table}
\endgroup

\textbf{The effect of training strategy}: CRN can not be well trained by adding the loss at the end or at each time step. We evaluate different training strategies on split 1 of HMDB-51 in Table. \ref{t:training-sample}, $a$ indicates supervising at the end, $b$ indicates supervising at each time step, $c$ indicates supervising at the end and one previous selected time step and $d$ indicates supervising at the end and two previous selected time steps. As stated in the introduction, $c$ balances the supervision strength within CRN and therefore, better performance can be achieved. Thus, $c$ is accepted for training CRN. 

\textbf{The effect of coupled recurrence}: From Table \ref{t:ucf101-single}, we can see that leveraging the reciprocal information of each other, both spatial and temporal networks can achieve better performance. Since our CRN needs a paired input and generates a paired output, spatial networks listed here are the average of the two CRNs which are trained by RGB images, flows and RGB images, RGB differences. The accuracy on split 1 of UCF-101 with a bninception backbone is $\textbf{90.4\%}$ for spatial networks, $\textbf{91.8\%}$ for temporal networks trained using flows and,  $\textbf{89.5\%}$ for temporal networks trained using RGB differences. With an inceptionv3 backbone we can achieve $\textbf{92.1\%}$ for spatial networks, $\textbf{93.5\%}$ for temporal networks trained using flows and,  $\textbf{91.6\%}$ for temporal networks trained using RGB differences on split 1 of UCF-101. \textit{They surpass all the independently trained two-stream networks in both spatial and temporal networks}. Although our N-CRN is not a strong model, the performance boosted by combining the models (CRN+N-CRN) is significant. Since statistics generated by the coupled training is different from independent training, we expect our CRN models to be a good compensation for any independently trained two-stream models. 

\begin{table}[htb] \caption{Evaluation on split 1 of UCF-101 and HMDB-51} \label{t:ucf101-single}
\scriptsize
\setlength{\tabcolsep}{2.7pt}
\centering
\begin{tabular}{c|c|c|c} \hline
\multicolumn{4}{c}{UCF-101} \\
\hline
Training setting&S-Nets&T-Nets (Flow) &T-Nets (Diff) \\
\hline
Clarifai \cite{simonyan2014twostream} &72.7\%  &81.0\% & - \\
VGGNet-16 \cite{linmin16tsn} &79.8\% &85.7\% & - \\
BN-Inception \cite{linmin16tsn}&84.5\% &87.2\% &83.8\%  \\
BN-Inception+TSN \cite{linmin16tsn} &85.7\% &87.9\% &86.5\% \\
\hline
N-CRN (bninception backbone) &84.7\% &85.6\% &86.2\%\\ 
CRN (bninception backbone)&90.4\% &91.8\% &89.5\%\\
CRN + N-CRN &91.0\% ($\uparrow$ 0.6)& 92.2\% ($\uparrow$ 0.4)&89.7\% ($\uparrow$ 0.2)\\
\hline
\hline
N-CRN (inceptionv3 backbone) &85.7\% &87.2\% &86.9\%\\ 
CRN (inceptionv3 backbone) &92.1\% &93.5\% &91.6\% \\
CRN + N-CRN &92.8\% ($\uparrow$ 0.7)&94.0\% ($\uparrow$ 0.5)&93.9\% ($\uparrow$ 2.3)\\
\hline
\multicolumn{4}{c}{HMDB-51} \\
\hline
Clarifai \cite{simonyan2014twostream} &40.5\%  &54.6\% & - \\
BN-Inception+TSN \cite{linmin16tsn} &54.4\% &62.4\% & - \\
\hline
N-CRN (bninception backbone) &51.4\% & 56.9\%&53.2\%\\ 
CRN (bninception backbone)&58.1\% & 67.5\%&59.0\%\\ 
CRN + N-CRN &59.0\%  ($\uparrow$ 0.9)&68.3\%  ($\uparrow$ 0.7)&60.7\%  ($\uparrow$ 1.7)\\
\hline
\hline
N-CRN (inceptionv3 backbone) &52.4\% & 57.9\%&54.9\%\\ 
CRN (inceptionv3 backbone) &62.7\% & 67.7\%&60.9\%\\ 
CRN + N-CRN &63.2\% ($\uparrow$ 0.5)&69.2\% ($\uparrow$ 1.5)&61.9\% ($\uparrow$ 1.0)\\
\hline
\end{tabular}
\end{table}

\begin{figure}[htp]
  \centering
  \centerline{\includegraphics[width=6.8cm, height=3.5cm]{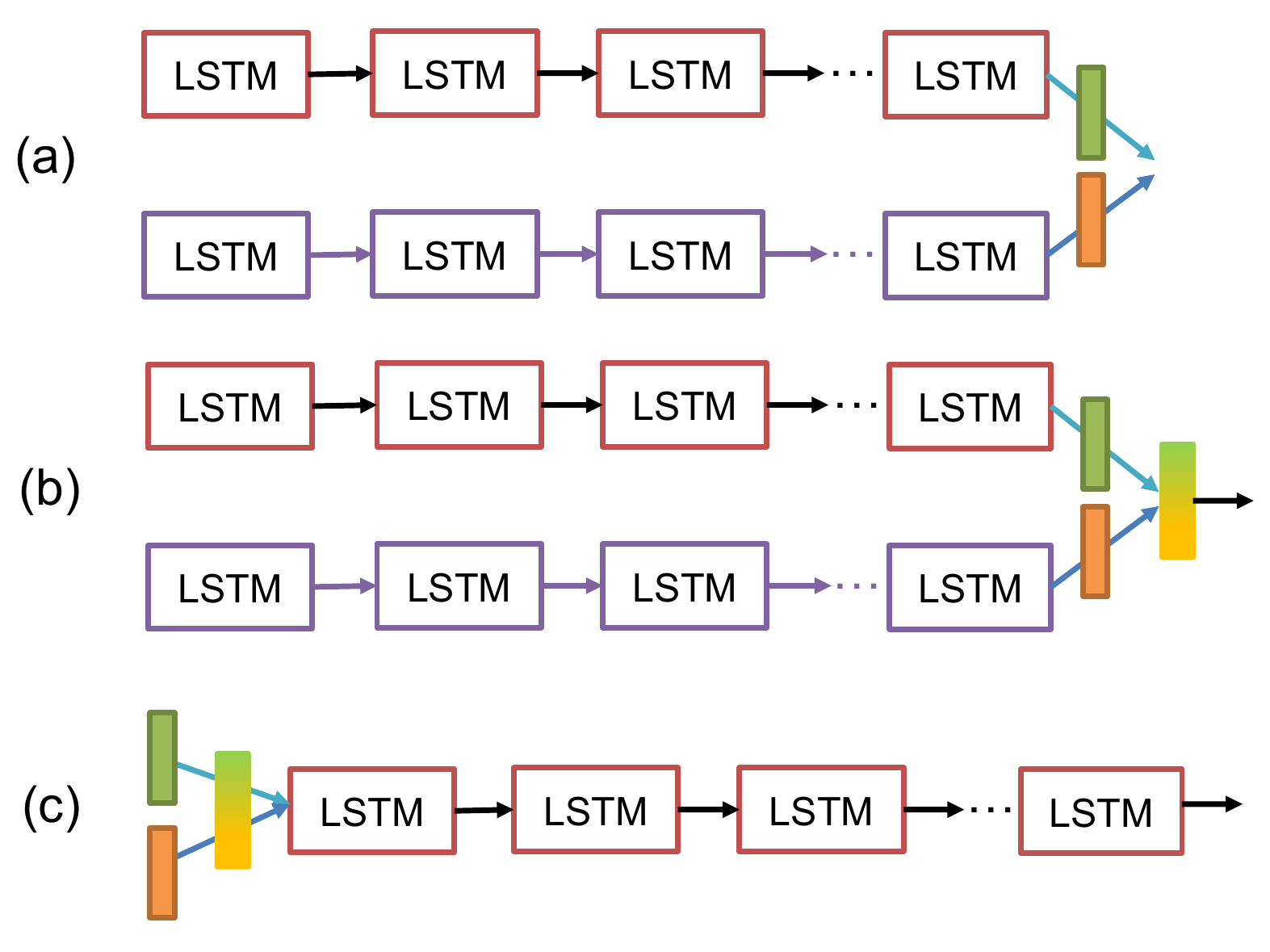}}
  \caption{Illustration of alternative designs for fusing different input sources. } \label{fig:alternative}
\end{figure}
\textbf{Comparison with alternative designs and other state-of-the-art methods}: The mean accuracy on the three splits of UCF-101 and HMDB-51 compared with the state-of-the-art and alternative designs can be seen from Table \ref{t:results}. CRN can achieve comparable if not better performance. Together with a N-CRN model, ours outperforms the state-of-the-art on both datasets by a large margin. We also present the alternative designs for the fusion of multiple input sources as shown in Fig. \ref{fig:alternative}. Most of previous state-of-the-art methods adopt (a), a two-stream architecture, we also experiment (b) and (c) for a fair comparison. (b) is a form of \textbf{late fusion}, two hidden outputs from two branches will be fed into a fusion module. (c) is a form of \textbf{early fusion}, the concatenated inputs will be fed into a fusion module and then pass through a recurrent network.  Even effective compared to some other methods, neither late fusion nor early fusion can provide better representations than CRN for action recognition task.

\begingroup
\setlength{\tabcolsep}{6.0pt}
\renewcommand{\arraystretch}{1}
\begin{table}[htp] \caption{Mean accuracy on the UCF-101 and HMDB-51 datasets} \label{t:results}
\scriptsize
\centering
\begin{threeparttable}
\begin{tabular}{c|c|c|c}
\hline
\multicolumn{2}{c|}{UCF-101} & \multicolumn{2}{c}{HMDB-51}     \\
\hline
\hline
EMV-CNN \cite{zhang16mv} &86.4 &EMV-CNN \cite{zhang16mv} & - \\
Two Stream \cite{simonyan2014twostream} &88.0 & Two Stream \cite{simonyan2014twostream} &59.4\\
$F_{ST}CN$ (SCI Fusion) \cite{sun2015fstcn} &88.1 & $F_{ST}CN$ (SCI Fusion) \cite{sun2015fstcn} &59.1\\
C3D (3 nets) \cite{du2015C3D} &85.2 & C3D (3 nets) \cite{du2015C3D} &-\\
Feature amplification \cite{park2016fusion} &89.1 &Feature amplification \cite{park2016fusion} & 54.9 \\
VideoLSTM\cite{li16videolstm} &89.2 &VideoLSTM\cite{li16videolstm} &56.4 \\
TDD+FV \cite{wang2015tdd} &90.3 &TDD+FV \cite{wang2015tdd} &63.2\\
Fusion \cite{feichtenhofer16fusion} &92.5 & Fusion \cite{feichtenhofer16fusion} &65.4 \\
$L^{2}STM$ \cite{lin2017l2stm}  & 93.6 &$L^{2}STM$ \cite{lin2017l2stm} & 66.2 \\
ST-ResNet \cite{feichtenhofer16resinet}   &93.4 &ST-ResNet \cite{feichtenhofer16resinet}  &66.4 \\
I3D \cite{i3d2017} & 93.4 &I3D \cite{i3d2017} & 66.4 \\
TSN \cite{linmin16tsn} &94.0 &TSN \cite{linmin16tsn} &68.5\\
Gated CNNs \cite{yudistira2017} & 94.1 & Gated CNNs  \cite{yudistira2017}  & 70.0 \\
\hline
Late fusion (bninception) & 92.4 & Late fusion (bninception) & 66.5\\
Early fusion (bninception) & 92.7 & Early fusion (bninception) & 66.3\\
\hline
N-CRN(bninception backbone)  & 92.2 & N-CRN(bninception backbone)  &65.7 \\
CRN(bninception backbone)   & 93.5 & CRN(bninception backbone)  &67.8 \\
CRN + N-CRN & \textbf{94.6} & CRN + N-CRN &\textbf{69.4} \\
\hline
\hline
Late fusion (inceptionv3) & 92.5 & Late fusion (inceptionv3) & 67.5\\
Early fusion (inceptionv3) & 92.7 & Early fusion (inceptionv3) & 67.1\\
\hline
N-CRN(inceptionv3 backbone)  & 93.1 & N-CRN(inceptionv3 backbone)  &66.3 \\
CRN(inceptionv3 backbone)   & 94.1 & CRN(inceptionv3 backbone)  &68.2 \\
CRN + N-CRN & \textbf{94.9} & CRN + N-CRN &\textbf{70.6} \\
\hline
\end{tabular}
\end{threeparttable}
\end{table}
\endgroup

Besides these relatively large datasets, we also evaluate CRN on a larger dataset, Moments in Time \cite{monfort2017moments}. The performance can be seen in Table \ref{t:moments}. CRN sets a new benchmark on Moments in Time by a lager margin. Single spatial and temporal model can beat the sophisticated assembled ones. 

\begin{table}[htb] \caption{Performance evaluation on Moments in Time} \label{t:moments}
\scriptsize
\setlength{\tabcolsep}{3.5pt}
\centering
\begin{tabular}{c|c|c|c} \hline
Model & Modality & Top-1 (\%) & Top-5 (\%) \\
\hline
Chance &- &0.29 &1.47 \\ 
ResNet50-scratch \cite{monfort2017moments} &Spatial &23.65 &46.73 \\
ResNet50-Places \cite{monfort2017moments} &Spatial &26.44 &50.56 \\
ResNet50-ImageNet \cite{monfort2017moments} &Spatial &27.16 &51.68 \\
TSN-Spatial \cite{monfort2017moments} &Spatial &24.11 &49.10 \\
CRN-Spatial  &Spatial & \textbf{27.32} & \textbf{50.01} \\
\hline
BNInception-Flow \cite{monfort2017moments} &Temporal &11.60 &27.40 \\
ResNet50-DyImg \cite{monfort2017moments} &Temporal &15.76 &35.69 \\
TSN-Flow \cite{monfort2017moments} &Temporal & 15.71 &34.65 \\
CRN-Flow &Temporal & 26.13 & 47.36 \\
CRN-RGBDiff  &Temporal  & \textbf{27.11} & \textbf{49.35} \\
\hline
TSN-2stream \cite{monfort2017moments} &Spatial+Temporal &25.32 &50.10 \\
TRN-Multiscale \cite{monfort2017moments} &Spatial+Temporal  &28.27 & 53.87 \\
Ensemble All \cite{monfort2017moments} & Spatial+Temporal + Audio & 30.40 & 55.94 \\
CRN + N-CRN &Spatial+Temporal &\textbf{35.87} & \textbf{64.05} \\
\hline
\end{tabular}
\end{table}

\subsubsection{Real Time Action Recognition}
Real-time action recognition is important for practical applications. Inspired by \cite{sun2015fstcn}, the RGB difference between the neighboring frames can be a good substitute of optical flows. Compared with optical flows which require certain amount of calculations, RGB difference can be directly inferred from RGB frames online without burden. Balance of speed and accuracy, RGB frames and RGB difference are the good input sources for real-time action recognition. However, as indicated in \cite{linmin16tsnj}, compared to optical flows, RGB difference only provides weak motion information which will degrade the performance. Since CRN enable spatial and temporal networks interpret complementary information from each other, the learning of RGBDiff can be calibrated by RGB, thus, we can achieve much better performance compared to previous state-of-the-art ($93\%$ vs. $91\%$). As shown in Table \ref{t:ucf101-realtime}, applying CRN boosts the real-time action recognition performance by \textbf{$2\%$}. 

\begin{table}[htb] \caption{Performance evaluation of real time action recognition} \label{t:ucf101-realtime}
\scriptsize
\setlength{\tabcolsep}{2.5pt}
\centering
\begin{threeparttable}
\begin{tabular}{c|c|c|c} \hline
Method           &Speed (GPU)	&UCF101 Split 1	&UCF101 Average\\
\hline
Enhanced MV \cite{zhang16mv}	&390 FPS	&86.6\%	&86.4\% \\
Two-stream 3Dnet \cite{ali2017e3D}	&246 FPS	&-	&90.2\% \\
RGB Diff w/o TSN \cite{linmin16tsnj}	&660FPS	&83.0\%	&N/A \\
RGB Diff + TSN \cite{linmin16tsnj}	&660FPS	&86.5\%	&87.7\% \\
RGB Diff + RGB (both TSN) \cite{linmin16tsnj}	&340 FPS	&90.7\%	&91.0\% \\
Ours (RGB Diff + RGB)	&200 FPS{*}	&\textbf{92.2}\%	&\textbf{93.0\%} \\
\hline
\end{tabular}
    \begin{tablenotes}
      \tiny
      \item {*} May vary when a different GPU is used.
    \end{tablenotes}
\end{threeparttable}
\end{table}

\subsection{Evaluation on Multi-person Pose Estimation}
\begin{figure}
  \centering
  \centerline{\includegraphics[width=7cm, height=4.5cm]{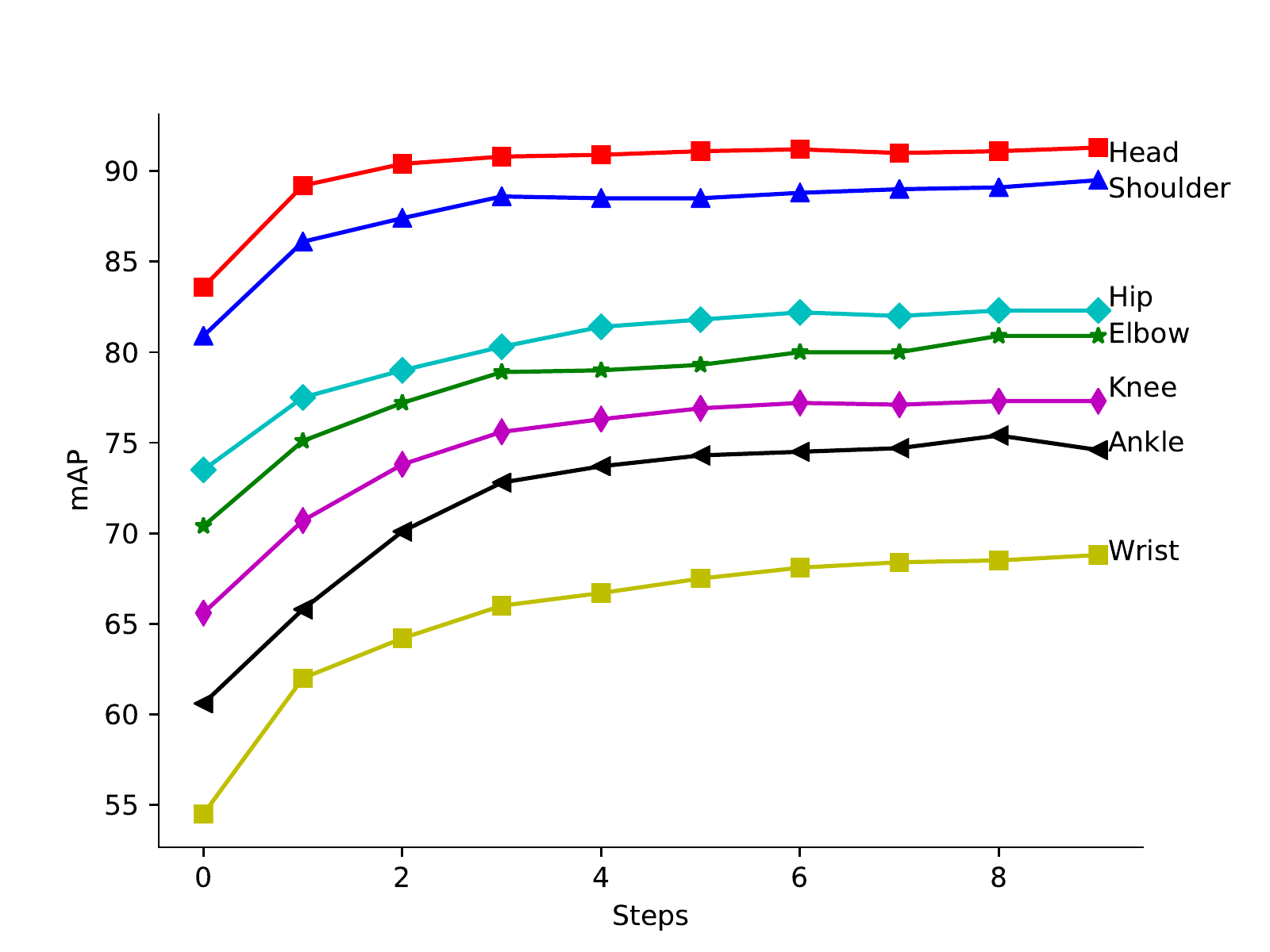}}
  \caption{Different joint location performance of MPII with the change of time steps. } \label{fig:map-steps}
\end{figure}
\vspace{-0.5em}
Human pose estimation is another dimension to analyze human activity. CRNs with different numbers of hidden maps is evaluated on MPII as shown in Table \ref{t:pose-val}. 
\begingroup
\setlength{\tabcolsep}{5.0pt}
\renewcommand{\arraystretch}{1.2}
\begin{table}[htp] \caption{Evaluation with different hidden features on MPII} \label{t:pose-val}
\scriptsize
\centering
\begin{tabular}{c c c c c c c c c} \hline
Arch & Hea & Sho & Elb & Wri & Hip & Knee  & Ank & \textbf{mAP} \\
\hline
%Cao et.al \cite{cao2016realtime} &91.2 & 89.4 & 78.7 & 65.7 & 80.8 & 75.6 & 71.6 & 79.0 \\
Cao et.al \cite{cao2016realtime} & 91.3  & 90.2  & 80.6  & 66.9  & 79.9  & 76.0 & 72.4 & 79.6 \\
$CRN_{F32}$  & 91.4  & 90.6  & 79.6  & 64.0  & 81.6  & 74.3 & 67.8 & 78.5 \\
$CRN_{F64}$  & \textbf{92.9}  & \textbf{91.4}  & \textbf{81.9}  & 69.4  & 82.8  & 77.8 & 73.4 & 81.4\\
$CRN_{F96}$  & 92.8  & 91.2  & \textbf{81.9}  & \textbf{69.9}  & 84.4  & 77.7 & \textbf{74.3} & 81.7\\
$CRN_{F128}$ & 92.1  & 90.1  & 81.0  & \textbf{69.9}  & \textbf{84.8}  & \textbf{80.3} & 74.2 & \textbf{81.8} \\
\hline
\end{tabular}
\end{table}
\endgroup

Here, $CRN_{F*}$ indicates a CRN with the corresponding `*' hidden feature maps in the LSTM. Even with 64 feature maps, CRN can exceed the state-of-the-art method \cite{cao2016realtime}. When the number of hidden feature maps increases, the performance becomes better. Note that even with 128 feature maps, the size of our proposed model is still smaller the model proposed in \cite{cao2016realtime}. 

\begin{figure}
  \centering
  \centerline{\includegraphics[width=9cm]{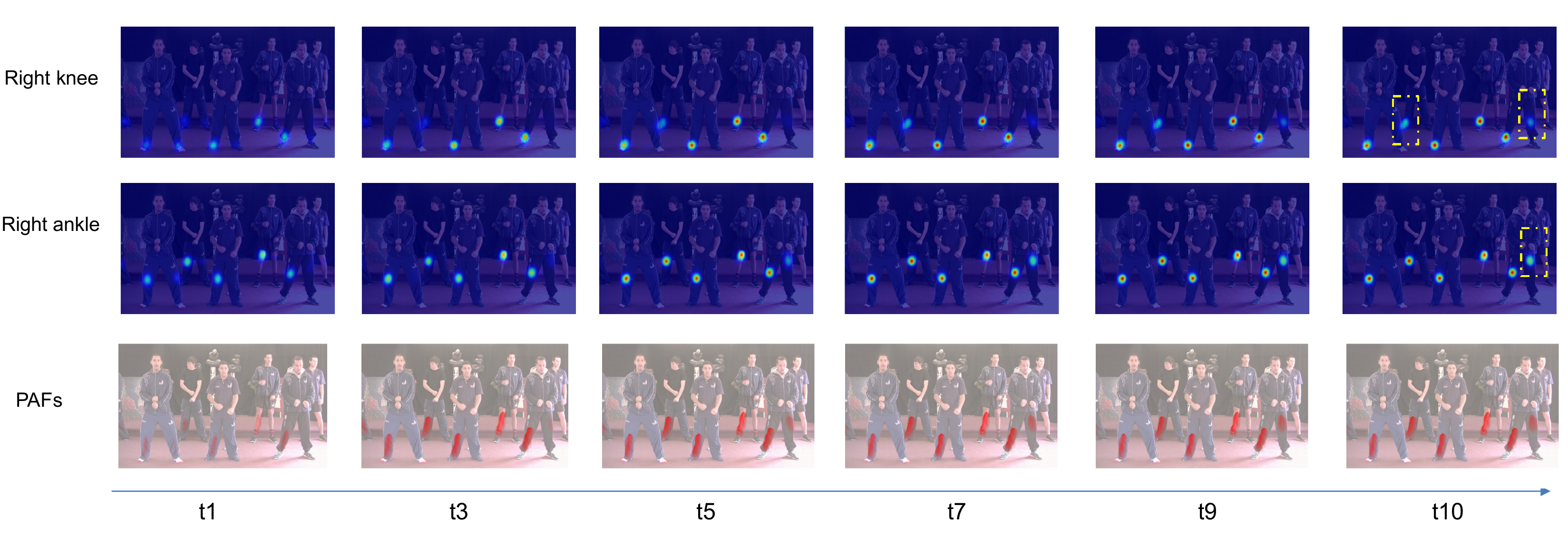}}
  \caption{The visualization of the heat maps and corresponding PAF at different steps. Better viewed in color and zoomed in.} \label{fig:vis-steps}
\end{figure}
The performance varies when different time steps are applied for the inference. As shown in Fig. \ref{fig:map-steps}, prediction of easy joint locations, such as, head or shoulders, even in early stage ($t \geq 3$), perform well. However, for more difficult ones, such as, wrist or ankle, more time steps are required for better performance. This experiments present the effectiveness of the recursive refinement for image based computer vision task using CRN. 

\subsection{Quantitative Performance Evaluation}
The joint location and PAFs prediction at different time steps are generated in Fig. \ref{fig:vis-steps}. As the time step increases, the joint as well as PAFs prediction becomes more and more confident (the brightness reveals the confidence level). Pay attention to the joint prediction shown in the dashed yellow rectangles along the time axis, initially, the confidence of prediction is pretty weak, however, with the time step increasing, the confidence is highly augmented. It verifies our assumption that CRN does iterative refining from one side. 

\begin{figure*}
  \centering
  \centerline{\includegraphics[width=14cm]{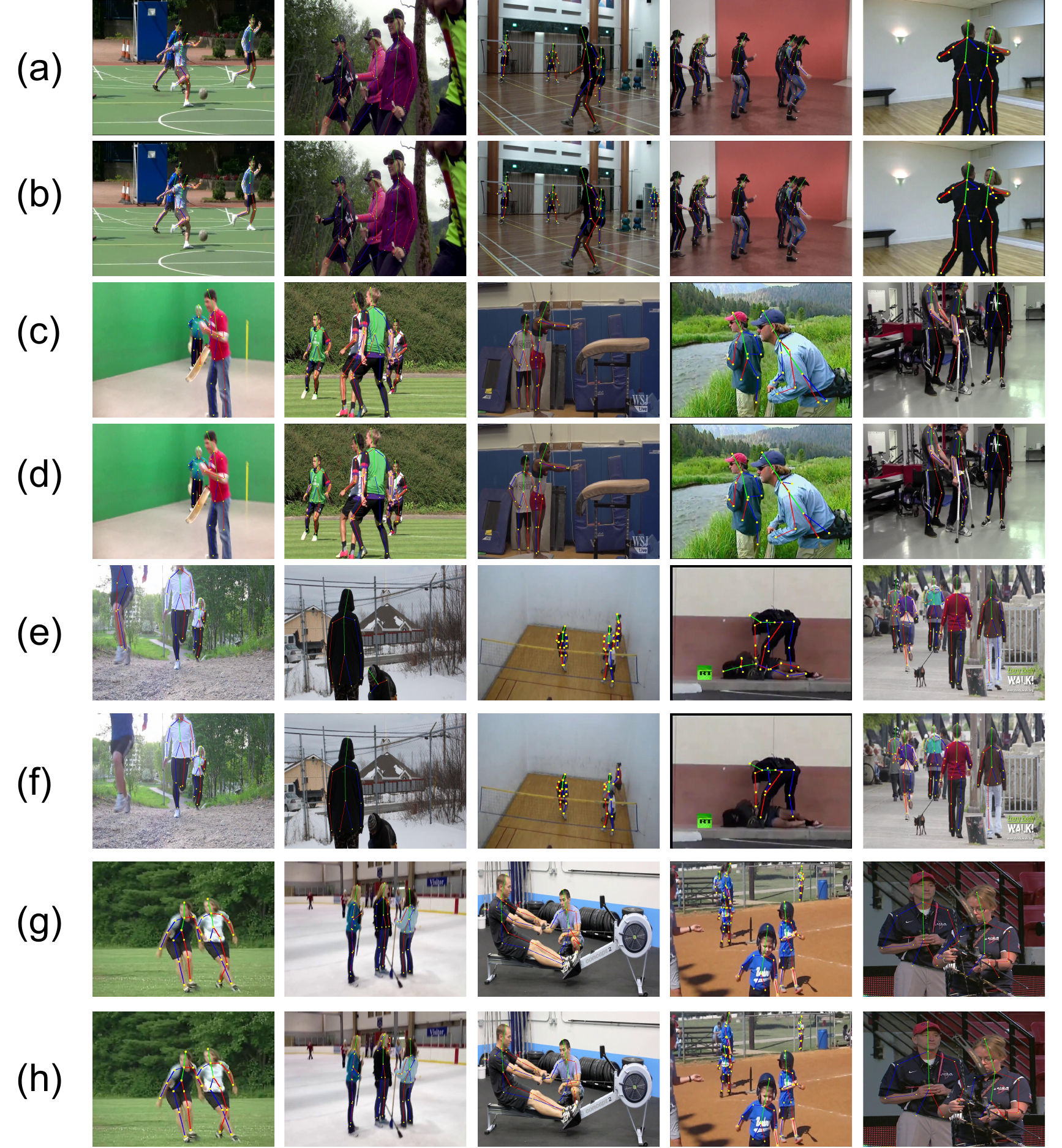}}
  \caption{The visualization of the pose estimation results on samples of MPII dataset. (a), (c), (e) and (g) are the results from CRN, (b), (d), (f) and (h) are the results from \cite{cao2016realtime}. Our proposed method presents better pose estimation in variation of viewpoint and appearance ((a) vs. (b), (e) vs. (f) and (g) vs. (h)) and occlusion ((c) vs. (d)). Better viewed in color and zoomed in. } \label{fig:compare}
\end{figure*}

The visualization of the pose estimation on sample images from MPII \cite{andriluka14cvpr} are shown in Fig. \ref{fig:compare}. 

In this figure, (a), (c), (e) and (g) are the results generated using CRN and (b), (d), (f) and (h) are the results generated using the method proposed in \cite{cao2016realtime}. From this figure, we can see that our proposed method, Coupled Recurrent Network (CRN), can deal well with rare poses or appearances with less/no false parts detection. Even for images with substantial overlap of the body parts of two people, our proposed method still works well, correctly associating parts for each person. (a) vs. (b), (e) vs. (f) and (g) vs. (h) presents that CRN can work well in different situations with variation of viewpoint and appearance. (c) vs. (d) shows that CRN can work better for occluded poses than other state-of-the-art method proposed in \cite{cao2016realtime}. 

\section{Summary} \label{SecConclude}
In this paper, we propose a novel architecture, called a Coupled Recurrent Network (CRN), to learn better representations from the multiple input sources. With the \textit{RIB} module, reciprocal information can be well distilled from the related input source. Iterative refinement using re-currency improves the performance step by step. Extensive experiments are conducted on two tasks, human action recognition and multi-person pose estimation. Due to the effective integration of features from different sources, our model can achieve the state-of-the-art performance on these human-centric computer vision tasks. Hope our work shed the light on other computer vision or machine learning tasks with multiple inputs. 

\section{Acknowledgment}

The author would like to thank Dr. Xingyu Zhang for constructive comments that greatly improved the manuscript.

{\small
\bibliographystyle{unsrt}
\bibliography{egbib}
}
\end{document}